# Enhancing Image Caption Generation Using Reinforcement Learning with Human Feedback

Adarsh N L[1], Dr. Arun P V[1], Aravindh N L[2]

[1] Indian Institute of Information Technology, Sri City.

[2] Shanmugha Arts, Science, Technology & Research Academy, Thanjavur.

adarsh.nl@iiits.in, arun.pv@iiits.in, 123004027@sastra.ac.in

*Abstract*—**Research on generative models to produce human-aligned / human-preferred outputs has seen significant recent contributions. Between text and image-generative models, we narrowed our focus to text-based generative models, particularly to produce captions for images that align with human preferences. In this research, we explored a potential method to amplify the performance of the Deep Neural Network Model to generate captions that are preferred by humans. This was achieved by integrating Supervised Learning and Reinforcement Learning with Human Feedback (RLHF) using the Flickr8k dataset. Also, a novel loss function that is capable of optimizing the model based on human feedback is introduced. In this paper, we provide a concise sketch of our approach and results, hoping to contribute to the ongoing advances in the field of human-aligned generative AI models.**

## I. INTRODUCTION

The process of automatically generating captions for images is called Image captioning. It has garnered significant attention in recent years, due to its applications in areas like Natural Language Processing (NLP), Computer Vision (CV), and Human-Computer Interaction (HCI). It has the potential to improve image interpretation, accessibility, and retrieval systems if captions for images can be generated in a precise and coherent manner. Automatically generating captions for images poses a significant challenge due to the inherent complexity involved in comprehending visual content and effectively transforming it into coherent natural language descriptions. Conventional methodologies depend on manual feature extraction techniques like Scale-Invariant Feature Transform (SIFT), Local Binary Patterns (LBP), and Histogram of Oriented Gradients (HOG) [27,28,29] combined with template-based caption generation methods [30]. Nevertheless, recent progress in deep learning, specifically in the domain of Convolutional Neural Networks (CNN), and Recurrent Neural Networks (RNN) has led to vital improvements in the understanding of landscape in images and quality of captions. An image encoder and a language decoder are the two main components of the architecture for generating captions for images [16]. Convolutional neural networks (CNNs) are used by the image encoder to extract meaningful visual features from the input image. The language decoder, on the other hand, employs recurrent neural networks (RNNs) to generate descriptive and contextually appropriate captions based on the encoded visual information. To train these models, large-scale image-caption datasets are used, where captions are paired with their corresponding images. Despite steady progress and breakthroughs in image captioning, there are still plenty of hurdles that must be overcome. These challenges include handling complex scenes, dealing with ambiguous or rare concepts, generating diverse and creative captions, generating human-preferred / human-like captions, and effectively incorporating context and world knowledge into the caption generation process. A coterie of independent researchers and organisations are actively swinging for the fences to further enhance the performance of the caption models by pushing boundaries and they set new milestones in this field. In the following article, firstly, we explore the area of image captioning by studying the state-of-the-art approaches and shedding light on their pros and cons. Following this, later, we also provide a novel loss function that has the potential to facilitate faster convergence of the models (refer figure 1). Through this work, we humbly contribute to the ongoing research in the field of human-aligned Generative AI models.

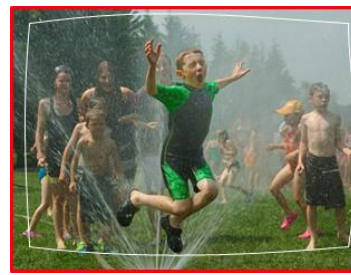

Kids playing with water in a grass field.

Fig. 1. Concept / Idea Figure. Our model generates descriptions for images by treating the captions as a label space.

## II. LITERATURE SURVEY

In 1995, Y. LeCun et al. [1] presented a comparison of different learning algorithms for the recognition and classification of handwritten digits. Then, in 2012, Krizhevsky et al., [2] revolutionized the world of image cognition and computer vision by introducing highly effective deep neural network (DNN) architecture for the ImageNet classification. The seed for everything that came after. Building on this, In 2013, Zeiler and Fergus, [3], focused on visualizing and understanding convolutional networks, providing valuable tools for interpreting the inner workings of deep neural networks.

Szegedy et al. proposed GoogLeNet [4], a deeper convolutional neural network with the introduction of inception modules, achieving state-of-the-art results in image classification. Simonyan and Zisserman, introduced the VGGNet architecture in [5], characterized by its simplicity and depth, which yielded outstanding performance in image classification tasks. Xu et al., [6], then presented deep residual learning, a groundbreaking concept that made it possible to train extremely deep neural networks effectively, mitigating the vanishing gradient problem. Building upon this work, Howard et al., introduced MobileNets in [7], an efficient convolutional neural network designed for mobile and embedded vision applications, enabling real-time deep learning on resource-constrained devices. Xu et al., [8], extended this into the domain of image captioning by proposing a neural model with visual attention, improving the generation of descriptive captions for images by aligning textual and visual elements. In parallel, Vinyals et al., combined convolutional and recurrent neural networks to generate image captions in [9], contributing to the field of neural image captioning. Meanwhile, Anderson et al., introduced attention mechanisms for image captioning and visual question answering [10], allowing models to focus on relevant image regions and improving performance on multimodal tasks.

Rennie et al., discussed self-critical sequence training for image captioning in [11], a novel technique for optimizing sequence-to-sequence models, and enhancing the quality of generated captions. In the domain of sequence generation, Vijayakumar et al., [13], introduced diverse beam search, a decoding strategy for generating diverse solutions from neural sequence models, enhancing the creativity of generated content.

Vedantam et al., [14], introduced the CIDEr metric, a consensus-based evaluation method for image descriptions, contributing to the development of more accurate and human-like captions. D. G. Lowe [15] dealt with the extraction of distinct image features from scale-invariant key points, an innovative advancement in the field of computer vision. Lu et al., then shifted focus to entity-aware image caption generation [16], enhancing image descriptions by considering entities present in the images, adding a new dimension to caption generation.

Gaurav and Pratistha Mathur, provided an overview of various deep learning models for automatic image captioning in [17], summarizing the state-of-the-art approaches in the field. Farhadi et al., [18], then went on to propose a method for generating sentences from images, pioneering the development of image captioning systems and multimodal understanding. Li et al., [19], discussed the composition of simple image descriptions using web-scale n-grams, a novel approach to generating image captions with large-scale language models. Kulkarni et al., introduced Babytalk in [20], a model for understanding and generating simple image descriptions, contributing to the advancement of image captioning research.

Zou et al., [21], introduced generalized decoding techniques for pixel, image, and language tasks, enhancing the generation of diverse and high-quality solutions in various domains. In parallel, Karpathy et al, [22,23], focused on deep visual-semantic alignments for generating image descriptions, emphasizing the alignment between images and natural language, facilitating more meaningful captions. To further improve our understanding of multimodal data, they also introduced deep fragment embeddings for bidirectional image-sentence mapping [23].

Kiros et al., [24, 25], combined visual-semantic embeddings with multimodal neural language models, enhancing the fusion of visual and textual data for understanding and generation. Expanding on this, in [25], they examined multimodal neural language models improving the understanding of text and images through their joint modelling. Finally, Kong et al., investigated text-to-image coreference in [26], addressing the complex task of connecting textual references to corresponding images in a multimodal context, broadening the realm of multimodal comprehension.

## III. PROPOSED WORK

We developed a Deep Neural Network architecture consisting of an image encoder and language decoder to generate captions for images. To improve the quality of the generated captions, we introduced a two-stage process: pre-finetuning and fine-tuning.

In the *pre-finetuning* stage, we manually evaluated and rated the captions generated by the model, to create a dataset that contains image features, model-generated captions, and the human rating for the generated captions. A *critic* model was trained with data to rate the captions, given the image and caption automatically.

In the *fine-tuning* stage, for each image being evaluated, we calculated the loss between the model-generated caption and the human-preferred caption along with the critic model rating and updated the model parameters by calculating the gradient to optimize the caption model performance. This fine-tuning approach allowed us to iteratively refine and align the base model's captions to human preference.

## IV. SYSTEM DESIGN

### A. The Base Caption Model

The base caption model is trained using the Flickr8k dataset. The dataset contains almost 8000 images and five captions describing each image. A few examples from the dataset and captions are shown in Figure 2. The preprocessing techniques employed on the data are detailed below.

*1) Data Preprocessing:* The first step involves loading the captions from a text file and organizing them into a dictionary (Key: image file name, values: set of captions). Each image is associated with a list of captions, allowing for easy access and retrieval during training. Subsequently, a cleaning procedure is applied to the captions to guarantee consistency and enhance the text data quality by converting the text to

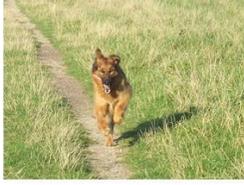

Dataset of images and captions

Training image:

Captions:
- A brown dog is running in a grassy plain .
- A brown dog runs along a path in the grass .
- Dog running in field
- Dog running on narrow dirt path .
- The dog is running along a path that has been made through the uncut grass .

Training image:

Captions:
- A boy in red shorts rides a yellow body-board on a breaking wave
- A boy surfs on a yellow board across the bright blue wave
- A surfer with a yellow board rides a wave
- Guy on yellow surfboard catching a wave
- Young man rides ocean wave on yellow surfboard .

Fig. 2. Sample images and their captions from the dataset.

lowercase, eliminating punctuations, and filtering out numbers. The cleaned captions are stored separately for later use.

*2) Vocabulary Creation and Reference Construction:* To facilitate a smooth evaluation of captions during training, a set of all unique words (Vocabulary) is created based on the cleaned captions. The diverse words in the vocabulary allow the model to generate a wide variety of captions. In addition, a reference set of captions is constructed by extracting all the captions from the cleaned dictionary and organising them into a list. This reference set is used for evaluating the BLEU score of the captions generated by the model.

*3) Feature Extraction using Transfer Learning:* Extracting features from the images is crucial in training the caption model. Figure 3 gives an abstract idea of how we are using transfer learning ¡add a reference to transfer learning¿ to extract features from the images. Pre-trained Xception model ¡add a reference to Xception model¿ trained for ImageNet classification (known for its excellent performance in image-related tasks) is used to extract the features from images. During feature extraction, the images are resized to meet the input requirements of the Xception model and they are fed to the model to get a compact representation of image features. These features are stored with their images and captions for subsequent use in training.

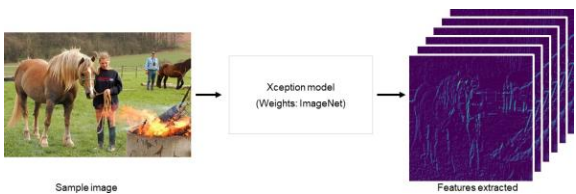

Sample image → Xception model (Weights: ImageNet) → Features extracted

Fig. 3. Feature extraction of images using the pre-trained Xception model.

*4) Data Preparation and Tokenization:* For each image, its feature (output of the Xception model) and corresponding captions are combined together to ensure that the data aligns correctly during training. A tokenizer is used to tokenize the captions (Takes the cleaned descriptions, converting them into a sequence of tokens). The tokenizer also assigns a numerical index for each unique word, which allows efficient representation and manipulation of text data during training.

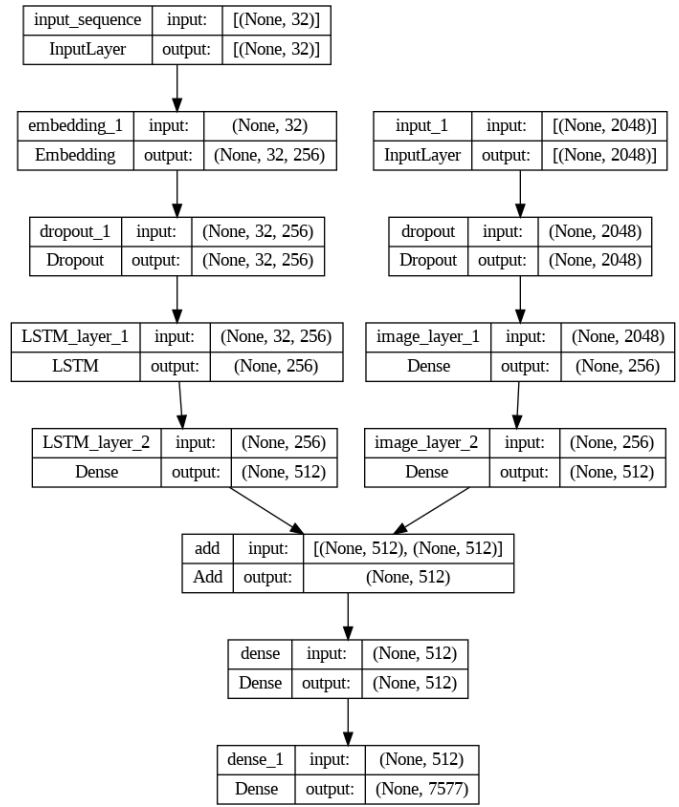

Fig. 4. Model architecture of the caption model

*5) Model Definition:* The model must be able to handle both the image features and tokenised sequences. The model architecture displayed in 4 is used to capture the relationship between image features and their corresponding tokenised captions. It learns the patterns and associations between them guided by optimizers like Adam and loss functions like categorical cross-entropy.

*6) Training model and Caption generation:* The training process involves iterating over the training data in batches. A specific function generates the training data by creating input-output pairs for the model. Each pair consists of image features and partial sequences of the captions as input, with the subsequent word in the captions as the corresponding output. During training, the model learns to predict the next word in the captions based on the given input. The training is performed for 10 epochs, with each epoch comprising a full pass through the training data. The train and validation loss of the model is displayed in Figure 5

Once the model is trained, the process of generating captions given an image involves a sequential and dynamic interplay between visual context and language modelling. Initially, high-level features are extracted from the input image using a pre-trained convolutional neural network, Xception as presented in Figure 3. These features serve as a foundation for caption generation. The caption generation process unfolds iteratively, where an initial token *start* is provided, and subsequent words are predicted sequentially. The evolving word sequence is tokenized, padded, and fed into a recurrent neural network, which refines its predictions based on learned patterns from training data. The incorporation of the image features ensures that the generated words are contextually relevant to the visual content. The sampling dynamics, controlled by a temperature parameter, introduce a level of randomness in word selection, influencing the diversity of predictions. The process continues until a predefined maximum length is reached, or the model predicts a termination token, typically *end* signalling the completion of the caption. This dynamic and context-aware approach enables the generation of descriptive and coherent captions that encapsulate the salient features of the input image.

evaluation process was recorded to facilitate the training of a special model known as *Critic* model. This critic model is intended to evaluate and rate the captions produced by the base caption model, combining human evaluators' insights into a quantitative framework that improves the model's captioning capabilities.

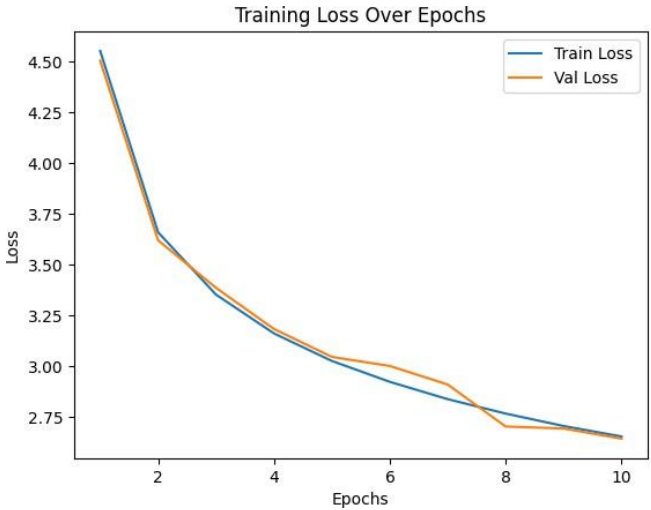

Fig. 6. Example captions predicted by our model, along with the human evaluator's feedback. Bounding boxes in the images are the feature landscapes, which we believe the model has considered for caption generation. For example in the first image (Left), the model has identified the man and horse correctly but failed miserably with grass and dogs. Thus the rating is -0.3. Similarly, for the second image(right), the model has understood the presence of a dog and grass in the image, which gives it a rating of 1.0.

### C. Stage 2: Fine tuning

In Stage 2: The fine-tuning process, the Reinforcement Learning with Human Feedback (RLHF) paradigm is brought into the picture for image captioning. The optimization process relies on a custom loss function (equation 1), encapsulating both traditional cross-entropy measuring dissimilarity between model-predicted and human-preferred captions, and feedback from the critic model, providing evaluative scores in the range of -1 to +1. This custom loss function is fundamental because of its ability to bridge the gap between model-generated captions and human-perceived relevance.

$$\text{Loss} = -\frac{1}{N} \sum_{i=1}^{N} \sum_{j=1}^{M} y_{ij} \cdot \log(p_{ij}) - \text{Feedback} \quad (1)$$

Here, $N$ represents the word count in the captions, $M$ is the vocabulary size, $y_{ij}$ represents the one-hot encoded vector for the $j$-th word in the $i$-th caption, and $p_{ij}$ denotes the predicted probability of the $j$-th word in the $i$-th caption.

The optimization of the model unfolds iteratively. For each image in the training set, the RL algorithm predicts a caption using the base caption model. The model's predictions are then compared with human-provided captions, and the custom loss is calculated. This loss along with *critic* model's feedback and the iterative refinement of the model's parameters, ensures a collaborative learning process that aligns the image captioning

Fig. 5. Train and Validation loss of the Base caption model.

### B. Stage 1: Pre - Finetuning

In the pre-finetuning(evaluation) phase, we conducted a manual assessment of the captions to create a *Critic* model. Initially, Utilizing images from the training set, we input them into the model to predict captions and human evaluators provided feedback on the relevance of each caption to the corresponding image. The evaluators assigned scores on a scale from -1 to +1, where -1 indicates poor and completely irrelevant captions, while +1 signifies good and entirely relevant captions. Figure 7 shows an example of human evaluation. This meticulous evaluation process was extended to cover all images in the validation set, and a subset of images from the training set was revisited for additional scrutiny. This model more closely with human-perceived quality and relevance. The entire process is carried out iteratively for a set of images, contributing to the continuous improvement of the captioning model.

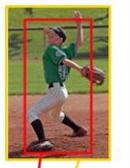

Fig. 7. Examples of generated captions, Human-preferred captions, and the Critic model's feedback. In the first, second and last image, the model has misunderstood the presence of a man/person but understood the landscapes like playfield, grass field and city correctly. The human feedback corrects the misunderstanding of the model here, as well as feeds their preference.

## V. RESULTS

This section covers the results we obtained from our research on enhancing image caption generation using reinforcement learning with human feedback.

The captions produced by the upgraded model are presented in Figure 8. An observable enhancement in caption quality is apparent through visual analysis, driven by human inputs. However, we acknowledge that there remains potential for further refinement in the model, particularly in the captions generated for certain images.

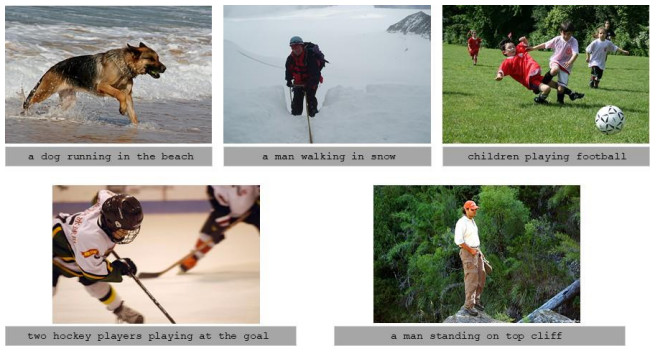

Fig. 8. Example descriptions generated by the enhanced model.

We evaluated the enhanced model's predicted captions with the base model based on the BLEU Score. We observed that the base model achieved a score of 9.19 and the enhanced model observed a score of 13.5.

## VI. CONCLUSION

In this paper, we outlined a multi-stage approach that integrated Supervised Learning and RLHF to enhance the quality of captions produced by the model, aiming for a more human-aligned output.

Our results indicate that using this approach proves successful in improving the quality of captions and our findings pave the way for future research, such as the incorporation of new evaluation metrics, diverse datasets, and the extension of the RLHF paradigm to different generative model architectures. In the future, we plan to further refine our techniques, and explore the applicability of our approach in various domains and datasets.